# Traffic map prediction using UNet based deep convolutional neural network

Sungbin Choi

sungbin.choi.1@gmail.com

**Abstract.** This paper describes our UNet based deep convolutional network approach on the Traffic4cast challenge 2019. Challenge's task is to predict future traffic flow volume, heading and speed on high resolution whole city map. We used UNet based deep convolutional neural network to train predictive model for the short term traffic forecast. On each convolution block, layers are densely connected with subsequent layers like a DenseNet. Trained and evaluated on the real world data set collected from three distinct cities in the world, our method achieved best performance in this challenge.

## 1. Introduction

In this paper, we describes our experiments on Traffic4cast (Traffic map movie forecasting) challenge 2019 [1]. Challenge's task is to predict short-term future traffic flow volume, heading and speed on high resolution whole city map. Dataset contains 100 billion probe points covering 3 cities (Berlin, Istanbul and Moscow) throughout a year. Real world road traffic conditions could be affected by various factors from hour to hour, day to day local event to a more long term seasonal effect. Given 1 hour's traffic data, we need to build model to predict next 15 minutes traffic condition. For a detailed explanation of challenge's task, please see challenge's website [1].

Since model should make prediction on every traffic map location, task's input and output has equivalent image size. UNet [2] has been used for these tasks, like image segmentation to label each pixel of image. We built model having UNet based architecture. Each convolution block is densely connected with subsequent layers like a DenseNet [3]. Our UNet based model could effectively predict future traffic conditions, achieving best performance in this year's challenge.

## 2. Methods

Each image frame contains aggregated traffic map information during 5 minutes time interval. We need to predict the next 3 images (equivalent to 15 minutes future) given past 12 images (equivalent to 60 minutes past). Firstly, sequence of input image is reshaped to (495, 436, 72) shaped data array (Section 2.1). Then UNet based model is applied to produce (495, 436, 9) shaped output (Section 2.3). Lastly, it is reshaped to (3, 495, 436, 3), having 3 channel image for each 3 future frame time point (Section 2.2).

### 2.1 Input

Original input is given as (12, 495, 436, 3) shaped data. Each dimension represents number of previous frame, height, width and channel. In last dimension, each channel characterize traffic volume, speed and direction. All channel's numeric values are scaled to lie within 0 and 255.

Traffic volume and speed channel represents continuous variable.

But direction channel indicates traffic direction category. In the original data, traffic directions are aggregated and categorized to one of five distinct category (northeast, northwest, southeast, southwest and none of above). Then arbitrary pixel values (0, 1, 85, 170 and 255 respectively) are assigned to it.

We split direction info into 4 separate discrete channels. For example, if traffic direction is labeled as 'northeast', output info is (1, 0, 0, 0). If traffic direction is labeled as 'none of above', output info is (0, 0, 0, 0).

So now input has shape of (12, 495, 436, 6).

In the first dimension, 12 frame sequence represents temporal information. In this study, recurrent model such as LSTM (long short-term memory network) [6] was not utilized. Temporal dimension is simply merged with feature channel (last dimension).

So now input has shape of (495, 436, 72).

### 2.2 Output

Prediction output should be (3, 495, 436, 3). Each dimension represents number of next frame, height, width and channel. Like 2.1, temporal dimension is merged with channel dimension. So now output from model has shape of (495, 436, 9).

### 2.3 Training

In this competition, there are 364 days data files per target city. There are 285 files for training, other 7 files used for validation. Each 1 day train file contains 288 (24 hour x 60 minutes / 5 minute interval) timeframe traffic map

Sliding window of timeframe is applied to extract input 12 timeframe image and subsequent 3 target output timeframe image, as described in Figure 1.

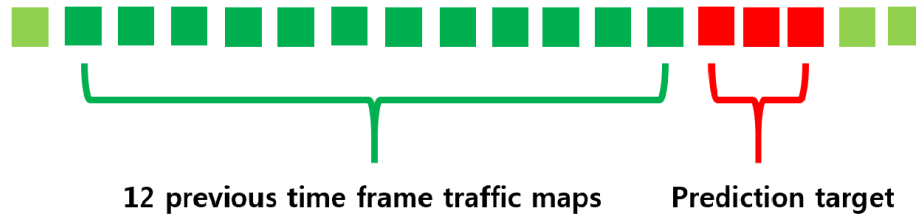

**Figure 1. Sliding window on time frames while training**

### 2.4 UNet based model

We implemented UNet [2] based model[1] on Tensorflow [4], as described in Figure 2 and Table 1. Each dense block consists of 4 convolutional layers, which are densely connected [3] (Figure 3). Mean squared error is used as loss function, with Adam optimizer [5]. Learning rate started from 3e-4 and manually lowered to 3e-5 when performance plateaued on validation set.

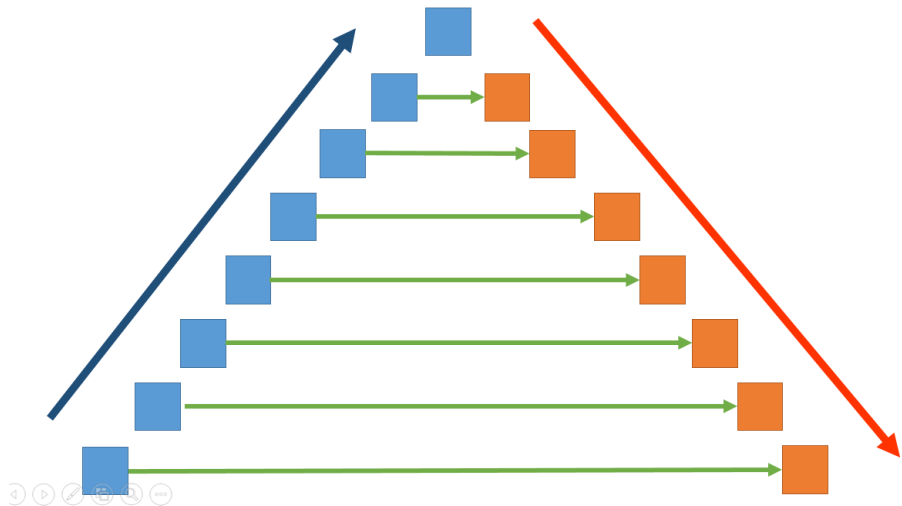

**Figure 2. Overall model structure**
*Each blue box represents dense convolution layers block with average pooling layer. Each orange box represents deconvolution layers. Green arrow represents skip connection between downsampling path and upsamping path.*

---

[1] Code is available online at https://github.com/sungbinchoi/traffic4cast2019

**2.5 Ensemble**

Building ensemble model [7] on top of distinct base neural network prediction might help to improve performance further. We trained four base model variants and added ensemble net on it. These four base models share same model structure described in previous section but have slight variations in the output shape or target loss function.

**Base model 1**
Same as previous section (2.4).
**Base model 2**
In the original input data, traffic map contains missing data region. In original input image, 'missing data' pixel values are simply set to zero. We masked out missing data pixel region from loss calculation to focus model prediction on 'have data' region.
**Base model 3**
Same as Base model 2, but this model trained on across all three cities.
**Base Model 4**
This model classifies each pixel into binary class as 'have data' or 'missing data'. Sigmoid cross entropy loss is used accordingly.

Output from each base model has been concatenated and used as input for ensemble. Ensemble model has same model architecture as base model in Section 2.3.

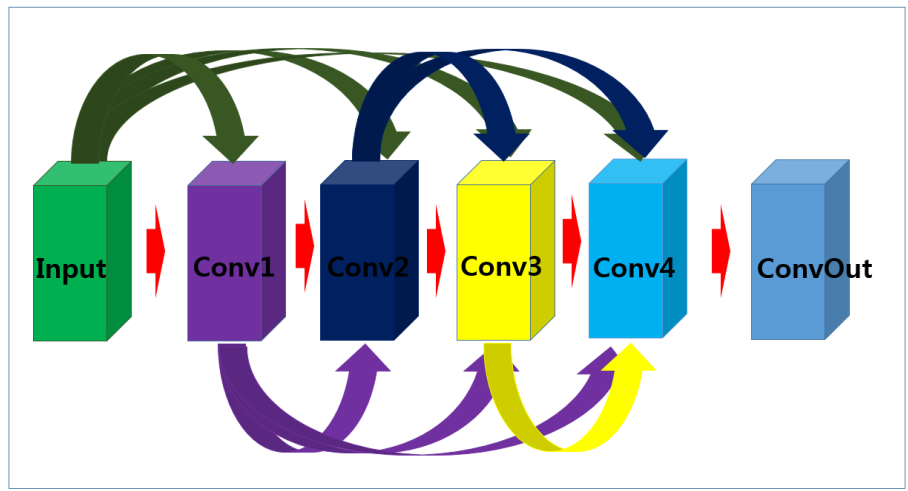

**Figure 3. Dense block structure**
*In our experimentation, 4 convolutional layers are densely connected with each other.*

## 3. Results

In this competition, there are 364 days data files per target city. 285 files used for training, 7 for validation, and 72 is used for held-out test set evaluation. Evaluation metric is mean squared error. Reference (gold standard) image data and predictions are normalized to between 0 and 1 before calculating mean squared error.

Table 1. Model output shape per each block

|  | Output shape |
|---|---|
| DenseBlock-1 | (495, 436, 64) |
| AveragePooling | (248, 218, 64) |
| DenseBlock-2 | (248, 218, 96) |
| AveragePooling | (124, 109, 96) |
| DenseBlock-3 | (124, 109, 128) |
| … |  |
| DenseBlock-7 | (8, 7, 128) |
| AveragePooling | (4, 4, 128) |
| DenseBlock-8 | (4, 4, 128) |
| Convolution Layer | (4, 4, 128) |
| DeconvolutionBlock-1 | (8, 7, 128) |
| DeconvolutionBlock-2 | (16, 14, 128) |
| … |  |
| DeconvolutionBlock-7 | (495, 436, 128) |
| Convolution Layer | (495, 436, 9) |

Our UNet based model started from height 495 width 436 sized input image and incrementally downsized using average pooling up to height 4 width 4 image at the top of the UNet (Table 1). Then it is deconvolutioned to produce back height 495 width 436 sized output image.

Table 2. Test set evaluation result

|  | Mean squared Error |
|---|---|
| All-zero baseline | 2.16957e-2 |
| Base Model 1 | 9.02919e-3 |
| Ensemble | 9.00773e-3 |

In Table 2, all-zero baseline represents prediction having all data values set to zero. Our UNet based deep convolutional neural network model showed high performance in this challenge outperforming other submitted methods.

Ensemble model draw additional performance gain but difference was very minimal. On ensemble model, mainly we tried to add missing data region prediction to optimize on the evaluation measure. But from the evaluation result, we suppose that missing data region might be randomly distributed or at least cannot be easily predicted in our models. So our ensemble approach was not very helpful in improving evaluation result.

## 4. Discussions

### 4.1 Optimal model structure

Every target task has different purpose and data characteristics. Initially, we experimented with various types of model to empirically find optimal model structure.

Fully convolutional network (FCN) [8] is a well known method in image segmentation task so we firstly experimented with it. Based on the same UNet architecture, simple convolutional network and ResNet [9] was also tried instead of DenseNet. All other models except ResNet showed inferior results. Due to computational burden of experimentation, we performed further experiments only with DenseNet throughout this challenge.

### 4.2 Augmenting training data with various techniques to fight overfitting

Image translation, rotation and image flipping techniques has been commonly used in imaging tasks. Augmenting small train data set by inducing small changes with data augmentation methods helps fight overfitting so makes model more generalizable in most cases.

We experimented with random image translation and flipping methods. In random image translation, image is randomly moved along x and y axis. In random image flipping, image is flipped horizontally or vertically, described as in Figure 4.

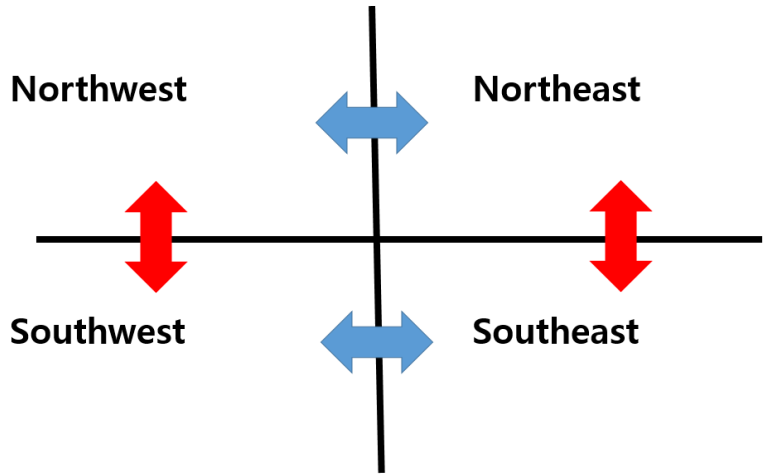

**Figure 4. Image flipping**
*Blue arrow represents horizontal flipping. Red arrow represents vertical flipping.*

In this challenge, input has three channels. Each channel represents traffic volume, speed and direction. We assumed that traffic volume and speed is neutral to the directional change. As described in Section 2.1, direction channel actually represents five distinct direction category (northeast, northwest, southeast, southwest and none of above). So when we flip the traffic map image, direction channel's category was set to change accordingly. For example, northwest direction is changed to southwest in vertical flipping. None of above category remains unchanged.

Contrary to our expectation, these image augmentation techniques actually did not improve or sometimes hurt performance. If original input traffic map image is not accurately aligned with north, south, east and west directional axis, image flipping can cause distortion of data. We also suspect that other possible cause might lie in the characteristics of direction channel data. Direction category information has been aggregated and averaged from numerical direction values captured from individual traffic objects, and then classified by 90 degree bin. Some of original directional information could be lost or distorted while aggregating. These questions could be found out when more details on input data acquisition and preprocessing methods revealed in the future.

## 5. Conclusion

In this study, we utilized UNet based deep convolutional neural network to predict future traffic flow volume, speed and direction on three major cities on earth. Each city represents separate local, geographic factors affecting traffic flow condition. UNet based method was effective to predict high resolution future traffic map. Real world application of these method can contribute to build accurate traffic forecast system or highly effective navigational system.